\pdfoutput=1

\documentclass[11pt]{article}

\usepackage{EMNLP2022}

\usepackage{times}
\usepackage{latexsym}
\usepackage{rotating}
\usepackage{lscape}

\usepackage[T1]{fontenc}

\usepackage[utf8]{inputenc}

\usepackage{microtype}
\usepackage{graphicx}
\usepackage{subcaption}
\usepackage{algpseudocode}
\usepackage{algorithm}

\usepackage{inconsolata}

\title{FastKASSIM: A Fast Tree Kernel-Based Syntactic Similarity Metric}

\author{Maximillian Chen\thanks{~~denotes equal contribution.},~
  Caitlyn Chen,$^{*}$~
  Xiao Yu,$^{*}$~
  Zhou Yu\\
  Department of Computer Science, Columbia University, New York, NY \\
  \texttt{maxchen@cs.columbia.edu} \\
  \texttt{\{caitlyn.chen, xy2437, zy2461\}@columbia.edu}
}

\begin{document}
\maketitle
\begin{abstract}
Syntax is a fundamental component of language, yet few metrics have been employed to capture syntactic similarity or coherence at the utterance- and document-level. The existing standard  document-level syntactic similarity metric is computationally expensive and performs inconsistently when faced with syntactically dissimilar documents. To address these challenges, we present FastKASSIM, a metric for utterance- and document-level syntactic similarity which pairs and averages the most similar constituency parse trees between a pair of documents based on tree kernels. FastKASSIM is more robust to syntactic dissimilarities and runs up to to 5.32 times faster than its predecessor over documents in the r/ChangeMyView corpus. FastKASSIM's improvements allow us to examine hypotheses in two settings with large documents. We find that syntactically similar arguments on r/ChangeMyView tend to be more persuasive, and that syntax is predictive of authorship attribution in the Australian High Court Judgment corpus.

\end{abstract}

\section{Introduction}
Syntax, the form of language, plays a crucial role in all aspects of natural language and communication, whether explicitly or implicitly. In storytelling, writers often have their own styles rooted in different syntactic tendencies \cite{feng2012characterizing}, allowing syntax to become indicators in prediction tasks such as gender~\cite{sarawgi2011gender} and authorship~\cite{raghavan2010authorship} attribution. Syntax also has social connotations in different cultures --- for example, in Russia, different social and demographic groups tend to use different syntactic patterns~\cite{bogdanova2016exploratory}. Such examples makes syntactic consistency crucial to capture in tasks such as machine translation and dialogue generation so that social conventions are not lost. Yet, recent research focuses primarily on evaluating similarity and coherence in terms of dimensions like semantics, the meaning behind language, (with approaches such as BERT embeddings~\cite{reimers-gurevych-2019-sentence,zhang2019bertscore}), or  lexical overlap (e.g., BLEU~\cite{papineni2002bleu}) --- even in work which uses syntax as an input to improve translation quality~\cite{zhang-etal-2019-syntax-enhanced}. 
\begin{figure}
    \centering
    \includegraphics[width=0.95\linewidth]{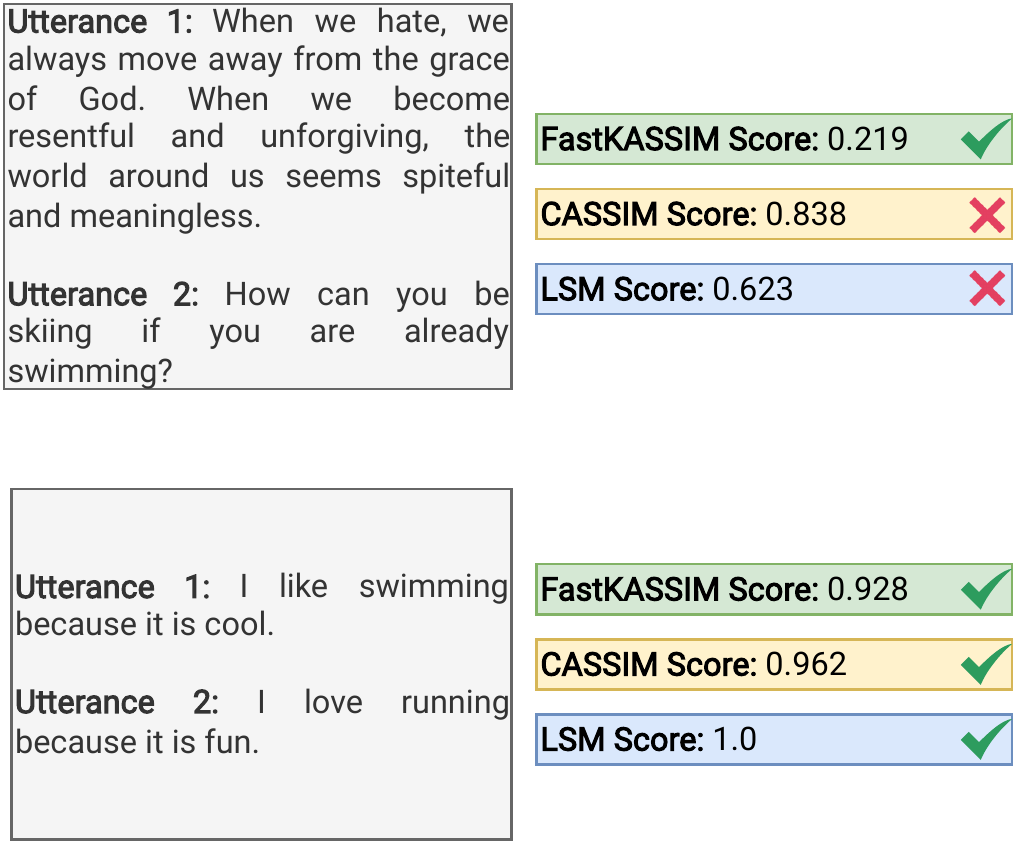}
    \caption{Comparison of FastKASSIM, CASSIM, and Linguistic Style Matching similarity scores. Top: two dissimilar utterances. Bottom: two similar utterances. All three metrics have strong agreement in cases of similar syntactic structure, but only FastKASSIM is able to recognize syntactically dissimilar utterances. The parse trees of these examples are visualized in Appendix~\ref{parse_trees}.}
    \label{fig:metric_comparison}
    \vspace{-4mm}
\end{figure}
A lack of work on syntax can be partially attributed to the absence of a practical, efficient metric that specifically compares syntax at the utterance level. The current standard metric is CASSIM~\cite{boghrati2016syntax, boghrati2018conversation}, but CASSIM uses a computationally expensive distance metric and can yield inconsistencies when comparing syntactically dissimilar documents (e.g., \autoref{fig:metric_comparison}).
To address this issue, we introduce \textit{FastKASSIM}, a \textit{Fast Tree Kernel-bAsed Syntactic SIMilarity Metric}\footnote{\url{https://github.com/jasonyux/FastKASSIM}},
 an improved metric for syntactic similarity at the utterance- and document-level. 

Like its predecessor, CASSIM, FastKASSIM computes the constituency parse tree of each sentence in a pair of documents, and the similarity between each pair of parse trees. But, while CASSIM used Edit Distance~\cite{pawlik2011rted,zhang1989simple} for similarity, we propose using a \textit{Label-based Tree Kernel} (henceforth LTK), our more syntactically thorough implementation of the Fast Tree Kernel~\cite{moschitti2006making}. We evaluate FastKASSIM against CASSIM and Linguistic Style Matching (henceforth LSM; ~\citet{niederhoffer2002linguistic, ireland2010language}). We find that FastKASSIM is more robust in cases of dissimilarities between documents and is generally more agreeable with human perception of differences in syntax. Additionally, the runtime of LTK is much faster than that of Edit Distance; it \textit{scales linearly with the number of node pairs with the same label} in a pair of parse trees. We empirically show large improvements in runtime with FastKASSIM. 

Previously, it was difficult to observe the role of syntax in behavioral phenomena at scale due to runtime constraints. Here, we contribute a study of hypotheses in two sets of applications. First, we examine the relationship between the persuasiveness of online arguments and syntactic similarity, and second, we observe the viability of syntax as an indicator in authorship attribution. FastKASSIM unlocks potential for evaluatory use in more contexts where it is important to preserve syntactic consistency and writing style, e.g., style transfer, machine translation, and story generation. 
\section{Related Work}
A few early studies focused solely on capturing syntactic structures. \citet{sagae2009clustering} sought to cluster \textit{words} by syntactic similarity. In order to establish a distance metric, they computed the cosine distance between vector representations of their unique constituency parses. Other approaches have used LIWC~\cite{tausczik2010psychological}. \citet{danescu2011mark} took a probabilistic approach to measure symmetry and influence of linguistic style.

Other early analytical work found that people will adjust their syntax to match dialog systems' syntactic~\cite{stoyanchev2009lexical} and lexical~\cite{stoyanchev2009lexical, hoshida2017lexical} choices. \citet{reitter2006priming} found that individual syntactic productions would repeat at low "distances" across utterances in both task-oriented and ``spontaneous'' dialog. For instance, \citet{reitter2007predicting} found that syntactic priming was predictive of success on the HCRC Map Task~\cite{anderson1991hcrc}. 
\citet{baker2021approaches} similarly discussed the use of syntactic similarity and overall linguistic style synchrony as an indicator of trust and cohesion in teamwork settings, and \citet{boncz2019communication} used syntactic similarity in modeling cognitive alignment. 

Syntactic features have been shown to improve prediction performance in downstream tasks, e.g., authorship attribution \cite{posadas2014complete, raghavan2010authorship,zhang-etal-2018-syntax} and gender attribution \cite{sarawgi2011gender}. In each case, the studies found significant performance gains from models that included syntax features.
Despite interest in syntax and clear improvements in prediction performance, the vast majority of recent work primarily focuses on semantic, or even lexical similarities/differences. This ranges from traditional methods such as TFIDF or Jaccard similarity to modern approaches including BERT embeddings~\cite{devlin2018bert, reimers-gurevych-2019-sentence,zhang2019bertscore} and AMR kernels~\cite{opitz2021weisfeiler}. Some approaches use syntactic features specific to certain domains such as Twitter~\cite{alnajran2019integrated,little2020semantic} or web documents~\cite{broder1997syntactic,pereira2003syntactic}. Other metrics include ``syntactic elements'' which take on various forms of parts-of-speech aggregation~\cite{alnajran2019integrated,pakray2011textual}.

A \textit{syntactic} similarity metric should appropriately consider differences in syntactic structure at the word-, utterance-, and document-level, as opposed to aggregating parts-of-speech or relying on domain-specific features. To our knowledge, CASSIM is the only metric to do this and has been proposed as a solution in applications ranging from measuring communicative alignment~\cite{boncz2019communication,baker2021approaches} to evaluating stylistic creativity in language learning~\cite{kokkola2022creativity} to clustering text~\cite{boghrati2017generalized}. However, CASSIM relies on the expensive Edit Distance metric, and occasionally assigns high similarity scores to documents that appear syntactically dissimilar. An improved syntactic similarity metric would afford new opportunities, from creating novel syntax feature vectors for classification tasks (e.g. authorship and gender attribution), to measuring syntactic coherence in machine translation. 
\begin{figure*}
    \centering
    \includegraphics[width=\linewidth]{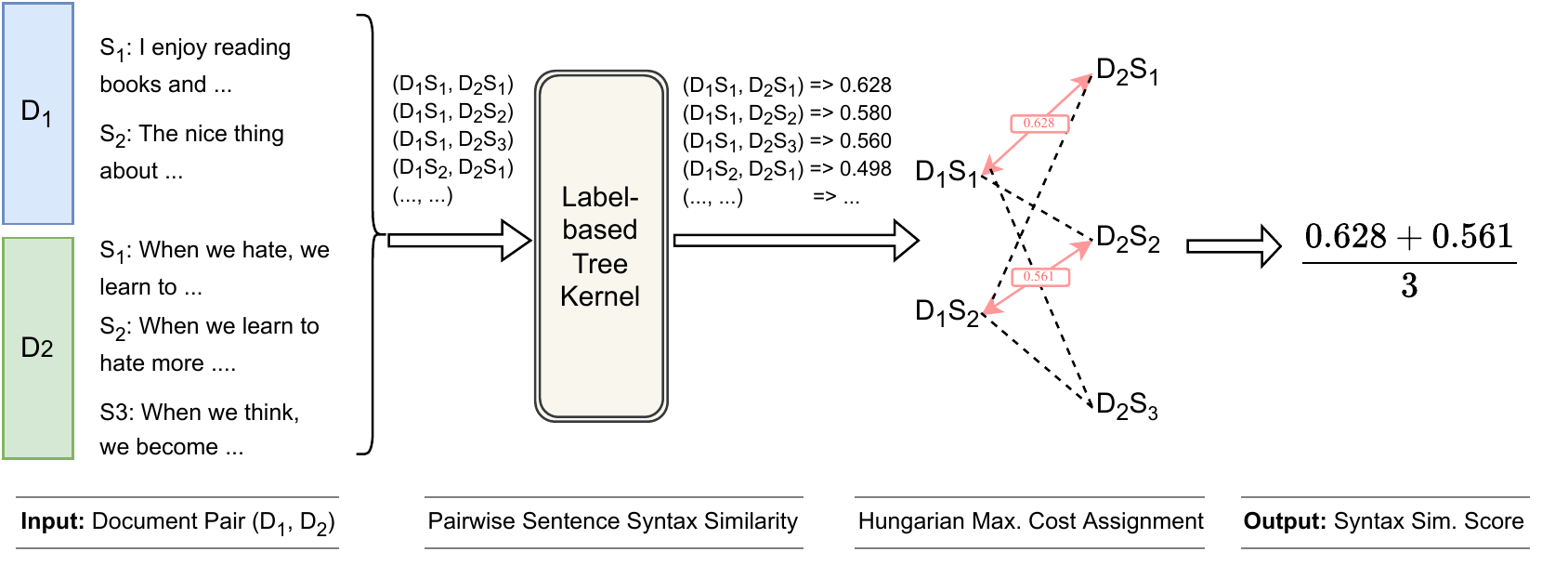}
    \caption{A high-level illustration of FastKASSIM computation. The parse trees of all sentence pairs between $D_1, D_2$ are computed using LTK. The Hungarian algorithm is used to pair together the most similar parse trees of each sentence in the two documents by the ``maximal cost'' (i.e., the largest tree kernels). The score is normalized by summing the paired kernel values then dividing by the number of sentences in the document with more sentences. $D_2S_3$ is unpaired because $D_1S_1, D_2S_1$ and $D_1S_2, D_2S_2$ are paired, and $D_2$ has more sentences than $D_1$.}
    \label{fig:fastkassim}
    \vspace{-4mm}
\end{figure*}
\begin{algorithm}
\caption{FastKASSIM}\label{alg:FastKASSIM}
\begin{algorithmic}[1]
\State DOCUMENTS $D_1, D_2$ 
\For{sentences $S_1, S_2$ in $D_1, D_2$}
    \State Compute Parse Tree($S_1$), Parse Tree($S_2$)
\EndFor
\For{Parse Trees $P_1, P_2$}
    \State Compute Tree Kernel:
    \State $s \gets 0$
    \For{Node Pair $n_1, n_2$ in $P_1, P_2$}
        \State $s \gets s + \Delta_{lb}$($n_1,n_2$)
    \EndFor
    \State Kernel $\gets$ normalize$(s)$
\EndFor
\State Hungarian Algorithm Max. Cost Assignment
\State \Return mean(maximal cost pairings)
\end{algorithmic}
\end{algorithm}
\section{FastKASSIM}
\subsection{CASSIM Background}
\label{sec:cassim_background}
CASSIM~\cite{boghrati2018conversation} was the first metric to compute syntactic similarity at the document-level. Their original algorithm uses the Stanford Parser~\cite{klein2003accurate, chen-manning-2014-fast} to compute the parse tree for each sentence in a pair of documents, before computing the  Edit Distance~\cite{wagner1974string} between each parse tree pairing. Then, they construct a bipartite graph and use the Hungarian Algorithm for minimum cost assignment~\cite{kuhn1955hungarian} to pair each tree in one document to the lowest distance tree in the second document. They finally average the Edit Distances of the minimal cost pairings. When there are different numbers of sentences, the number of assignments will correspond to the number of sentences in the document with fewer sentences. Each of that document's sentences will get paired with the most similar sentence in the second document, and the least similar sentences in the second document will remain unpaired. The final Edit Distance between a pair of parse trees $P_1, P_2$ is normalized as $\frac{Edit Distance}{Size(P_1) + Size(P_2) - 2}$, where $Size(P)$ is the number of nodes in $P$.

An important advantage of CASSIM is that it is generalizable to any corpus; it does not represent syntax using platform-specific features like~\citet{alnajran2019integrated,little2020semantic}. However, the cost of exhaustively using a metric such as Edit Distance is rather penalizing, as its implementations range in asymptotic time complexity from $\Theta(mn)$~\cite{wagner1974string} to $O(s \times min(m,n))$~\cite{ukkonen1985algorithms}, where $m$ and $n$ are the string lengths, and $s$ is the maximal Edit Distance. 
\begin{algorithm}
\caption{Delta$_{lb}$ Function ($\Delta_{lb}$)}\label{alg:delta}
\begin{algorithmic}[1]
\State Tree Nodes $n_1, n_2$; $cache$
\State Decay $\lambda$; Subtree/Subset Tree Indicator $\sigma$
  \If{$n_1,n_2$ is cached}
    \State \Return cache$\left(n_1,n_2\right)$
  \EndIf
  \If{$n_1,n_2$ have different labels}
    \State \Return $0$
  \EndIf
  \If{both $n_1,n_2$ are preterminals}
    \State cache$(n_1,n_2) \gets \lambda$ if same label, else $0$
    \State \Return cache$(n_1,n_2)$
  \EndIf
  \State Product $\gets 1$
  \For{child $c_1$ of node $n_1$}
    \State Accumulator $\gets 0$
    \For{child $c_2$ of node $n_2$}
      \State  Acc. $\gets$ Acc. $+$ Delta$_{lb}$($c_1,c_2$)
    \EndFor
    \State Product $\gets$ Product $\times (\sigma +$ Acc.$)$
  \EndFor
  \State cache$(n_1,n_2) \gets \lambda \times $ Product
  \State \Return $\lambda \times $ Product
\end{algorithmic}
\end{algorithm}
\subsection{The FastKASSIM Algorithm}
In large multi-sentence documents, repeated Edit Distance becomes the most expensive component of CASSIM. Thus, we propose FastKASSIM, which avoids the expensive Edit Distance computation by using Tree Kernels~\cite{moschitti2006making}. Tree Kernels can greatly reduce time complexity by caching between recursive subcalls. The Fast Tree Kernel algorithm \citet{moschitti2006making} runs in \textit{linear time on average with respect to parse tree sizes}. 

\begin{figure*}
    \centering
    \includegraphics[width=\linewidth]{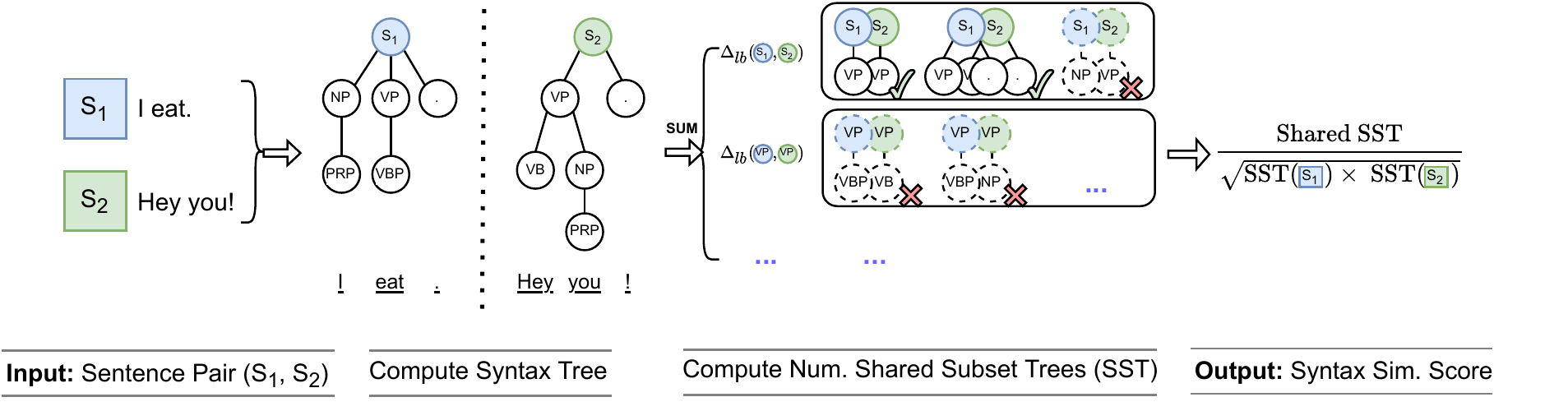}
    \caption{Overview of the Label-based Tree Kernel. The parse trees of a pair of sentences are computed, along with the number of common fragments rooted in each pair of parse tree nodes (i.e., number of shared subset trees). This is normalized by dividing by the square root of the product of the number of subset trees in each parse tree. }
    \label{fig:ltk}
    \vspace{-4mm}
\end{figure*}

We propose FastKASSIM, which replaces CASSIM's Edit Distance with a new normalized Tree Kernel. \autoref{fig:fastkassim} provides a high-level overview of FastKASSIM, which is formally described in Algorithm~\ref{alg:FastKASSIM}.
However, the Fast Tree Kernel does not allow for the case in which two parse tree nodes have matching labels but different productions. We thus also introduce the Label-based Tree Kernel (LTK)\footnote{There is a very strong correlation between LTK and the Fast Tree Kernel ($R=0.97$, $p$<0.001).}, which compares the labels at each node in a pair of subtrees or subset trees\footnote{\citet{moschitti2006making} defines a subtree as a node and all its descendants, whereas a subset tree does not require its leaves to be terminal.}. This also more closely follows~\cite{collins2002new}, which proposes comparing the actual subset trees rooted at two nodes in each parse tree, rather than the production. Figure~\ref{fig:ltk} depicts the LTK algorithm computing the number of shared subset trees in a pair of parse trees. More formally, as described in lines 7-11 of Algorithm~\ref{alg:FastKASSIM}, LTK accumulates the value of $\Delta_{lb}$ (Algorithm~\ref{alg:delta}), which is the number of common fragments rooted in a pair of parse tree nodes $n_1, n_2$. 

We follow \citet{moschitti2006making} by normalizing $LTK(T1, T2)$ as $\frac{LTK(T1, T2)}{\sqrt{LTK(T1, T1) \times LTK(T2, T2)}}$. This normalized tree kernel is not biased towards tree shape, in contrast to CASSIM's normalized Edit Distance ($\frac{Edit Distance}{Size(P_1) + Size(P_2) - 2}$). Under CASSIM's normalization, if two sentences resulted in the \textit{same parse tree size despite being composed of entirely different labels}, the normalized Edit Distance approaches 0.50 (the expression approximates $\frac{Size(P_1)}{2 \times Size(P_1)-2}$). In other words, according to CASSIM, sentences with the same shape but different parts-of-speech should be neither similar nor dissimilar. We further highlight possible examples of this bias by visualizing parse trees in Appendix~\ref{parse_trees}. Ultimately, our normalized LTK results in significant runtime improvements over Edit Distance, as we show in Section~\ref{runtime_comparison} and derive in Appendix~\ref{runtime_analysis}, and agrees strongly with human perception, as we show in Section~\ref{evaluation}. 

Like the original CASSIM algorithm, our high-level algorithm allows for flexibility in the choice of which parser to use, allowing for future improvements in runtime and correctness as research in parsing progresses. FastKASSIM similarly allows for flexibility in the implementation of tree kernels. Our implementation will be publicly released upon acceptance. In order to directly compare FastKASSIM and CASSIM, we default to using the Stanford Parser~\cite{chen-manning-2014-fast} and LTK, our aforementioned modified approach to the Fast Tree Kernel~\cite{moschitti2006making}.\footnote{However, we provide users with a native interface to interchangeably use any parser supported by NLTK~\cite{bird2009natural}.}
\subsection{Overall Metric Runtime Comparison}
\label{runtime_comparison}
The largest difference in the runtime of CASSIM and FastKASSIM is that CASSIM uses a
normalized Edit Distance to evaluate parse tree similarity, while FastKASSIM uses a normalized tree kernel. 

LTK recursively computes $\mathrm{\Delta_{lb}}$  across all $n_1, n_2$ pairs in parse trees $P_1, P_2$. 
But, importantly, \textit{all comparisons are cached to avoid repetition.} This results in LTK having an asymptotic runtime complexity of $O(S_{12})$, where $S_{12}$ is the total number of pairs of nodes in a pair of parse trees $P_1$, $P_2$ that have the same label. We prove this runtime in Appendix~\ref{runtime_analysis}. This is a large improvement over Edit Distance's runtime complexity of $O(s \times min(m,n))$. We confirmed that these asymptotic improvements apply to real-world uses cases by comparing how Edit Distance and LTK scale with the product of parse tree sizes in Figure~\ref{fig:rt_nm_cmp} of the Appendix, finding that LTK scales sublinearly while Edit Distance scales superlinearly.

\begin{figure}[t]
    \centering
    \includegraphics[width=\linewidth]{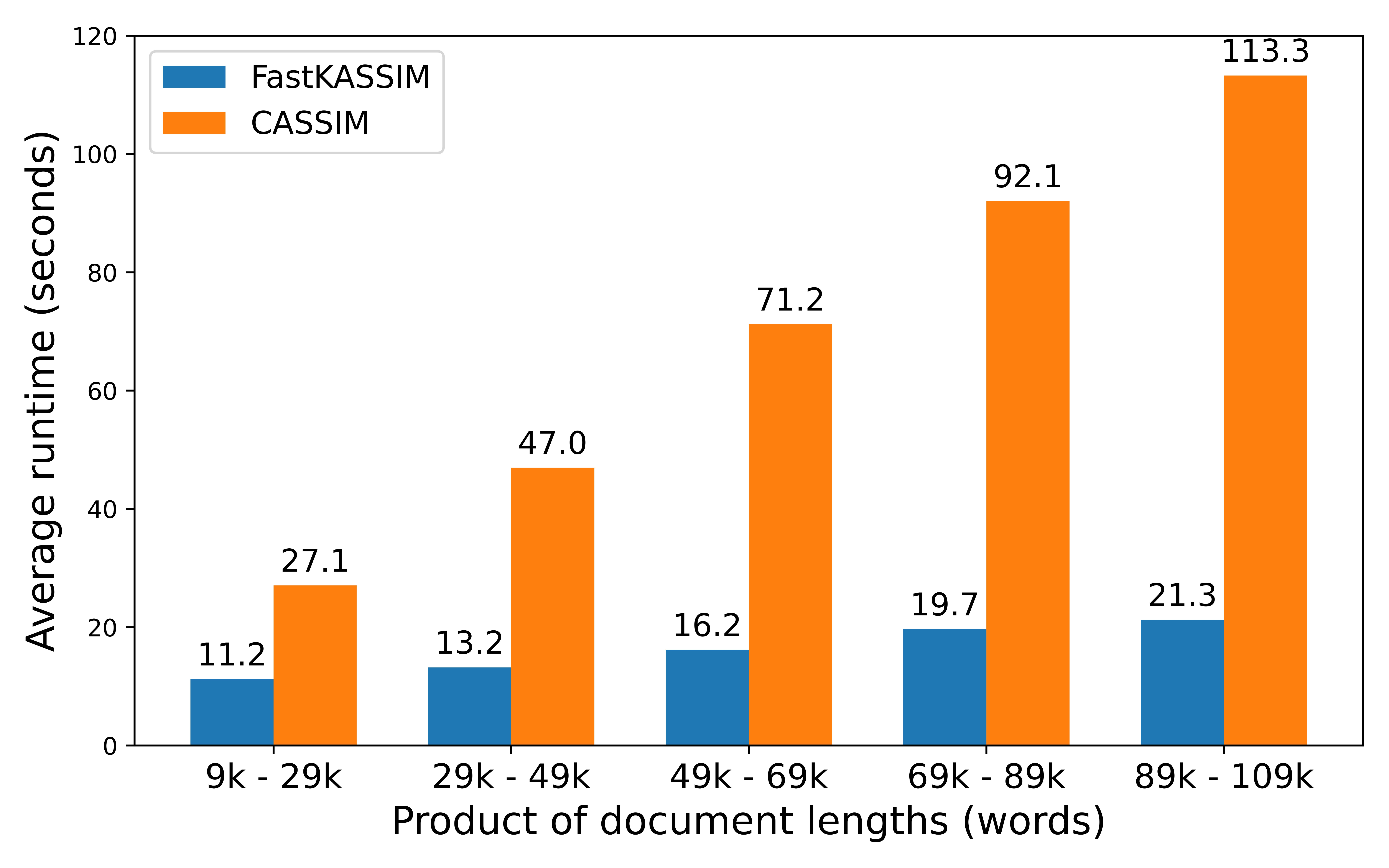}
    \caption{Runtime comparison between FastKASSIM and CASSIM. On the CMV corpus, FastKASSIM runs 2.42 times to 5.32 times faster on average, depending on document size.}
    \label{fig:full_rt_cmp}
    \vspace{-4mm}
\end{figure}

While Figure~\ref{fig:rt_nm_cmp} indicates that LTK can be up to an order of magnitude faster than Edit Distance, the largest bottleneck in overall time is still the time to compute each parse tree. Thus, in Figure~\ref{fig:full_rt_cmp}, we investigated the difference in "end-to-end" runtime between FastKASSIM and CASSIM without precomputing the parse trees. 

In this experiment, the ChangeMyView dataset (henceforth CMV; \citet{tan2016winning}) is used, providing a corpus of unstructured text with large document sizes, to evaluate the promises of FastKASSIM and CASSIM for their abilities to process entire documents. First, we sample entire document pairs and record the time it takes to compute the syntactic similarity of each pair. 
Each pair is randomly sampled from the 18,363 posts in the CMV training set.  
Then, we exhaustively paired documents based on the product of their document sizes, providing an approximation of the number of comparisons between parse trees. The document length for each CMV root posts has high variance, so document length products are grouped into bins. For each bin, we randomly sample 60 document pairs and report the average runtime.

\autoref{fig:full_rt_cmp} shows that FastKASSIM scales well in
runtime as the product of document lengths increases. For instance, when syntactic similarity between documents of lengths 300 words and 310 words were compared (product of $93,000$), CASSIM needed on average $113.3$ seconds while FastKASSIM took only $21.3$ seconds on average.
Given these drastic improvements in time complexity, it is now more feasible to compute syntactic similarity at the document level for large corpora. 
\section{Evaluating FastKASSIM}
\label{evaluation}
In this section, we first demonstrate FastKASSIM's overall ability to differentiate between similar and dissimilar documents. Then, we correlate its scores with CASSIM and LSM. Finally, we discuss FastKASSIM's advantages by explaining discrepancies in scoring.
\subsection{Discriminating Between Syntactically Similar and Dissimilar Documents}
\citet{boghrati2018conversation} validated CASSIM by comparing whether it was consistent with human perception of syntactic similarity. The authors asked Mechanical Turkers to write syntactically similar sentences given a sentence prompt. This resulted in a dataset of 472 English documents from 118 anonymous human annotators, and the authors found that CASSIM was able to ``identify syntactically similar documents.'' 

Following \citet{boghrati2018conversation}, we computed the syntactic similarity between each pairing of sentences generated both within the same prompt and between different prompts. Each individual prompt is structurally quite different (Table~\ref{prompts}).
By construction, \textit{documents resulting from the same prompt should be syntactically similar, whereas documents resulting from different prompts should be dissimilar}. As per their work, we fit a maximal structure linear mixed effect model with an indicator for whether the sentence corresponded to the same prompt or a different prompt as a fixed effect and the document ID as a random effect against standardized syntactic similarity\footnote{We use $z$-score standardization, $\frac{x - \mu}{\sigma}$.}.

We computed ANOVA of this full model against the reduced model, which drops the comparison type indicator as a fixed effect. The ANOVA $\chi^2$ test on the impact of comparison type yields statistically significant differences in the distribution of FastKASSIM ($\chi^2=1438.9$, $p$ $<0.001$), CASSIM ($\chi^2=331.84$, $p$ $<0.001$), and LSM ($\chi^2=201.85$, $p$ $<0.001$) scores between syntactically similar and dissimilar documents. FastKASSIM results in the largest effect size, $1438.9$, indicating it creates the largest differences in distribution. 
\subsection{Correlating Syntax Metrics}
\begin{table}[]
\small
\begin{tabular}{p{\linewidth}}
\textbf{Prompts} \\ \hline
1. The two most important days in your life are the day you are born and the day you find out why. The nice thing about being a celebrity is that you bore people and they think it’s their fault. \\ \hline 
2. I am enough of an artist to draw freely upon my imagination. Imagination is more important than knowledge. Knowledge is limited. Imagination encircles the world. \vspace{1pt}                               \\ \hline 
3. When we love, we always strive to become better than we are. When we strive to become better than we are, everything around us becomes better too.                                               \\ \hline
4. What is the point of being alive if you don’t at least try to do something remarkable?  
\end{tabular}
\caption{Prompts for the crowd-sourced corpus collected by \citet{boghrati2018conversation}.}
\label{prompts}
\vspace{-4mm}
\end{table}
LSM\footnote{We compute LSM using an implementation publicly available at https://github.com/miserman/lingmatch.} is a metric computing similarities from function word categories. This has ties to matching syntax, as those matching function words correspond to specific parts-of-speech. Moreover, LSM is a widely accepted metric for synchrony and correspondence of general linguistic style in documents (e.g. \citet{chartrand2005beyond,ludwig2013more}).
We examine the actual similarity scores calculated in the previous section on the crowdsourced document similarity corpus collected by \citet{boghrati2018conversation} using each of LSM, FastKASSIM, and CASSIM. In Figure~\ref{fig:corr_scatterplots}, we see that there is a moderately strong correlation between FastKASSIM and LSM ($R = 0.5$, $p$ $< 0.001$). This indicates that FastKASSIM is able to detect matches in key parts-of-speech. We would not expect to see a greatly higher correlation, because LSM is a measure of function words rather than a holistic measure of syntax. On the other hand, while we see a statistically significant correlation between CASSIM and LSM, its correlation coefficient is much lower ($R=0.11$, $p$ $< 0.001$), indicating a smaller connection between CASSIM representations and function words. This is likely due to the biased Edit Distance normalization mentioned in Section~\ref{sec:cassim_background}. Moreover, we actually find an overall negative correlation between FastKASSIM and CASSIM ($R=-0.33$, $p$ $<0.001$) with a seemingly bipartite relationship. There is an apparent disagreement over documents that FastKASSIM deems dissimilar, with agreement over documents that FastKASSIM deems similar. 
\begin{table}[]
\centering
\small
\begin{tabular}{l|l|l|l|l|l}
Metric       & Acc. & SR   & SP   & DR   & DP    \\ \hline
LSM        & 46.2 & 92.5 & 30.8 & 30.7 & 92.5  \\
LSM$_a$      & 65.6 & 81.1 & 40.6 & 60.4 & 90.6  \\
CASSIM     & 25.1 & \textbf{100.} & 25.0 & 0.11 & \textbf{100.}   \\
CASSIM$_a$   & 48.8 & 47.7 & 23.8 & 49.2 & 73.8  \\
BERTScore & 25.0 & 100. & 25.0 & 00.0 & 00.0 \\
BERTScore$_a$ & 74.6 & 99.3 & 49.6 & 66.4 & 99.6 \\
Sentence-BERT & 18.9 & 19.8 & 74.0 & 2.70 & 0.20 \\
Sentence-BERT$_a$ & 34.3 & 9.50 & 19.2 & 59.3 & 39.3 \\
FastKASSIM & \textbf{88.3} & 96.1 & \textbf{69.1} & \textbf{98.5} & 85.6 
\end{tabular}
\caption{Evaluation of LSM, CASSIM, BERTScore, Sentence-BERT and FastKASSIM in terms of Accuracy (Acc.), Similar Document Recall (SR), Similar Document Precision (SP), Dissimilar Document Recall (DR), and Dissimilar Document Precision (DP). Metric$_a$ denotes adjusting to a uniform distribution by quantile transformation.}
\label{metric_performance}
\vspace{-4mm}
\end{table}
\subsection{Discrepancies Across Syntax Metrics}
\label{discrepancies}
\begin{figure*}[t]
    \centering
      \begin{subfigure}{0.33\textwidth}
        \includegraphics[width=\linewidth]{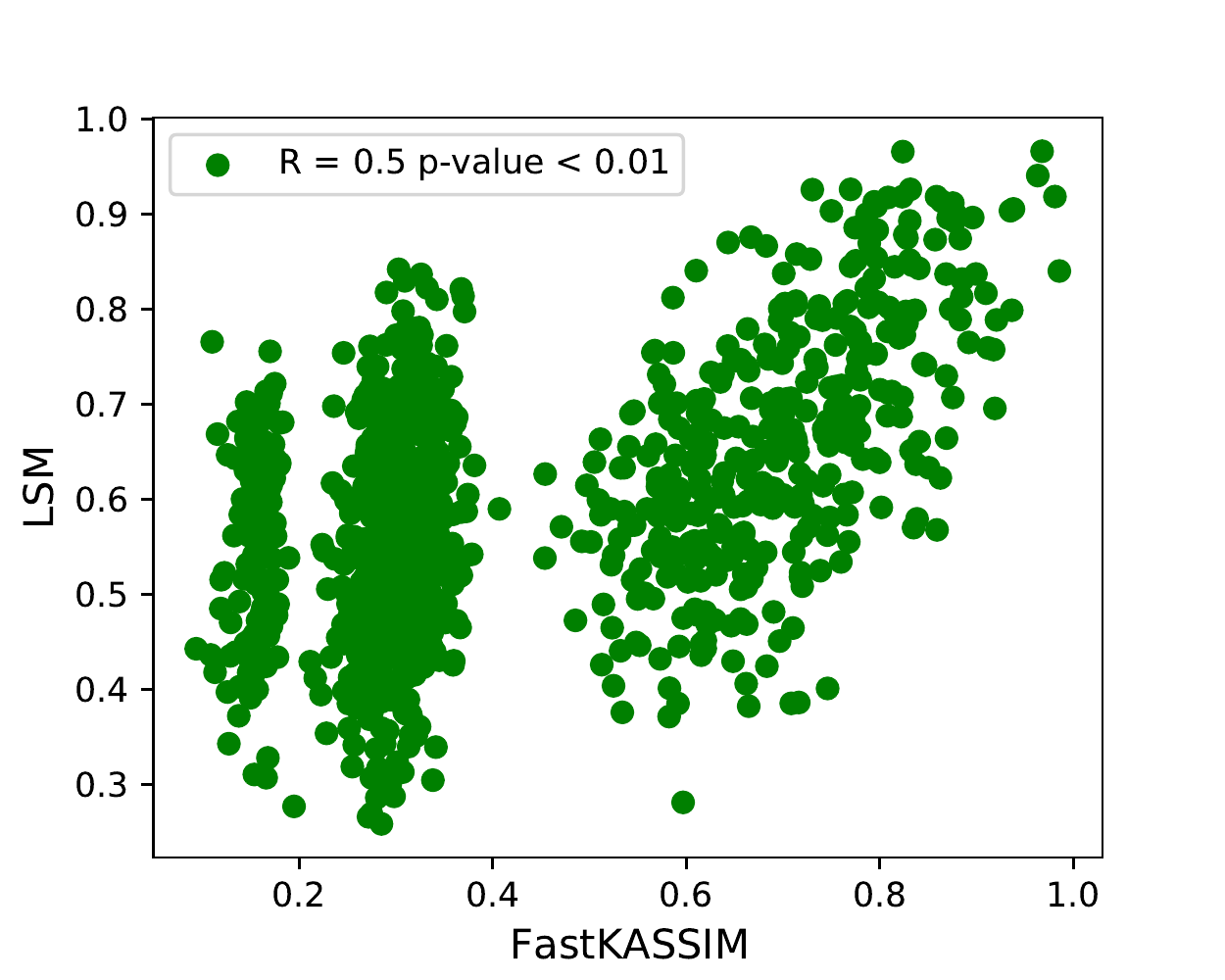}
        \caption{} \label{fig:ftklsm}
      \end{subfigure}%
      \hspace*{\fill}   
      \begin{subfigure}{0.33\textwidth}
        \includegraphics[width=\linewidth]{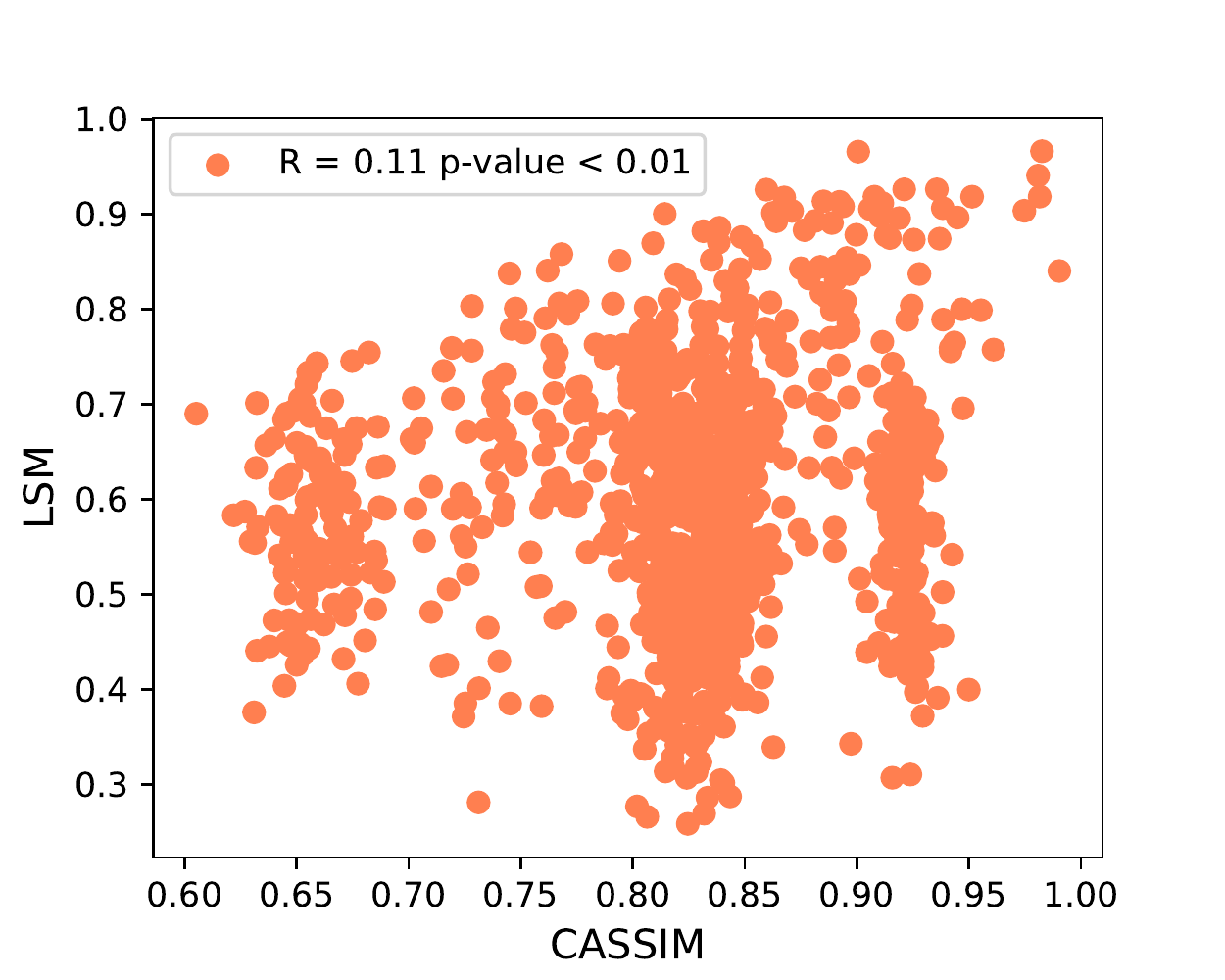}
        \caption{} \label{fig:cassimlsm}
      \end{subfigure}%
      \hspace*{\fill}   
      \begin{subfigure}{0.33\textwidth}
        \includegraphics[width=\linewidth]{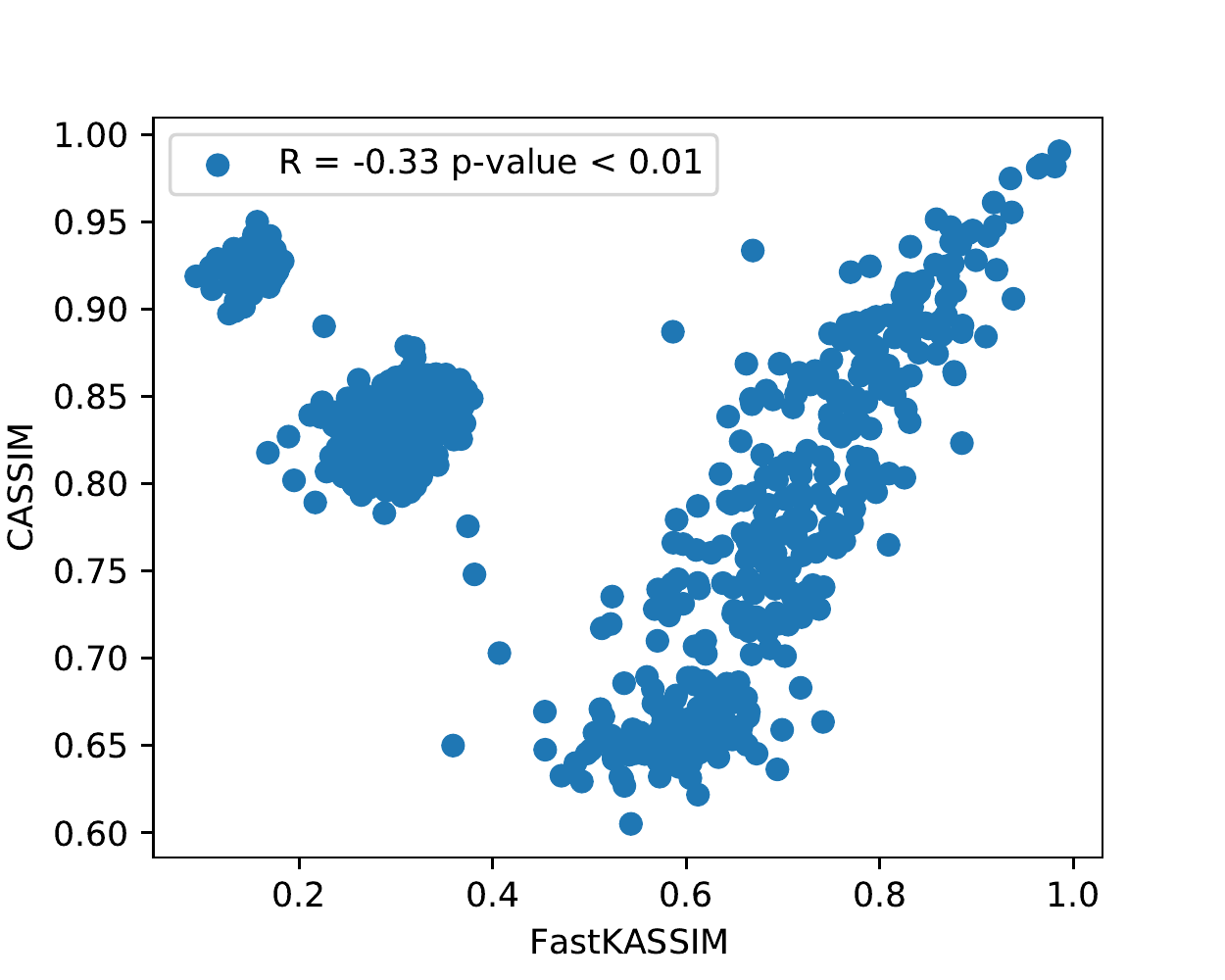}
        \caption{} \label{fig:ftkcassim}
      \end{subfigure}

    \caption{(a). FastKASSIM v. LSM. Moderately strong positive correlation with $R = 0.5$ and $p$ $< 0.01$. (b). CASSIM v. LSM. Weak but statistically significant positive correlation with $R = 0.11$ and $p$ < 0.01. (c). FastKASSIM v. CASSIM. Moderately strong negative correlation with $R = -0.33$, $p$ $< 0.01$.}
    \label{fig:corr_scatterplots}
    \vspace{-4mm}
\end{figure*}
Figure~\ref{fig:corr_scatterplots} indicates that there are several regions of disagreement (vertical clustering). In one region of Figure~\ref{fig:ftklsm}, FastKASSIM assigned low scores (less than 0.4) despite LSM ranging from $0.258$ to $0.842$. Recall that in this corpus, every utterance resulting from the same prompt was perceived as syntactically similar, and utterances from two different prompts were perceived as syntactically dissimilar. We find that in 248 out of 249 cases where FastKASSIM assigned a score below 0.4 yet LSM assigned a high score (above 0.6), the two documents being compared came from different prompts. 
 This implies that FastKASSIM indeed is discriminating between different syntactic structures, whereas LSM may be picking up on similarities other than syntax, as expected.

Figure~\ref{fig:cassimlsm} does not indicate any obvious relationship between CASSIM and LSM. More surprisingly, Figure~\ref{fig:ftkcassim} shows that for document pairings that FastKASSIM deems syntactically dissimilar (values of less than 0.5), there is a very strong negative correlation between FastKASSIM and CASSIM ($R=-0.783$, $p$ $<0.001$). Visually, there is a cluster of pairings where FastKASSIM assigns a value less than 0.4 but CASSIM assigns a value larger than 0.75. We find that in 677 of the 678 pairings in this cluster, the documents originate from different prompts, indicating that the documents are syntactically dissimilar. 

We evaluate these discrepancies in Table~\ref{metric_performance} in terms of each metric's ability to correctly identify documents originating from the same prompt as syntactically similar and those from different prompts as syntatically dissimilar using accuracy, recall, and precision\footnote{Exact expressions provided in Appendix~\ref{metric_equations}.}. In addition to CASSIM and LSM, we include an evaluation of BERTScore~\cite{zhang2019bertscore} and Sentence-BERT~\cite{reimers-gurevych-2019-sentence}. While they are not syntax metrics, they are strong embedding-based metric which may account for syntactic forms.

As similarity scores lie between 0.0 and 1.0, we use 0.5 as a boundary between similar and dissimilar documents. We reason that in unknown contexts, 0.5 is neutral.
We also compare against quantile transformations of each baseline metric, which map each metric's scores to a uniform distribution (note this is an unfair advantage, since in real-time deployment, one cannot observe the entire distribution of values), due to their apparent biases in Figure~\ref{fig:corr_scatterplots}.
In Table~\ref{metric_performance}, we find that FastKASSIM holistically outperforms both CASSIM and LSM, with the exception of similar document recall and dissimilar document precision. In these cases, CASSIM achieves perfect precision and recall because it only classified one document pair as dissimilar. BERTScore similarly yields a small range in similarity scores on this corpus. After undergoing a quantile transformation, BERTScore's sensitivity to syntactic differences is magnified, and it performs well in the aforementioned categories but underperforms FastKASSIM in accuracy, similarity precision and dissimilarity recall.

As indicated by its low Dissimilar Document Recall in Table~\ref{metric_performance}, we found CASSIM frequently assigned high similarity scores to syntactically different documents. This likely comes from the bias in its normalized Edit Distance as discussed in Section~\ref{sec:cassim_background}, and we show the significant improvements achieved by FastKASSIM.

 Overall, our results indicate that with respect to human intuition of syntax, FastKASSIM is more robust than CASSIM, LSM and BERTScore. In Appendix~\ref{parse_trees}, we visualize comparisons of several pairs of parse trees along with their FastKASSIM and CASSIM scores.

\vspace{-2mm}
\section{Applications}
\vspace{-2mm}
FastKASSIM is a more accurate and efficient syntactic similarity metric than the current state-of-the-art, opening the possibility for investigating new hypotheses in data-heavy fields with large corpora. Existing applications use syntax metrics for classification (e.g. \citet{posadas2014complete}) as well as analytically to measure hypotheses (e.g. \citet{kaster2021global}). Here, we use syntax as a linguistic style indicator in authorship attribution, and measure syntactic similarity to study communication accommodation in persuasive arguments. 
\subsection{Persuasiveness of Syntax Accommodation}
\label{sec:persuasion}
Early work in communication accommodation theory found that matching communication styles can create a sense of familiarity, which improves social and conversational outcomes (e.g. \citet{curhan2007thin,giles2016communication}). While most existing work has focused on hypotheses at the word-level~\cite{tan2016winning}, we hypothesize that CMV arguments that are more syntactically similar to opinions may be more persuasive as well.\footnote{Appendix~\ref{sec:persuasion_appendix} includes full details on CMV and preprocessing.}  

On CMV, users write an original post describing an opinion and allow ``challengers'' to present arguments attempting to change their opinion. Original posters (OP) indicate whether their opinions have been changed by assigning a ``delta,'' which we can use an indicator of successful persuasion. 
We computed the syntactic similarity between a challenger's initial argument and an original poster's (OP) original opinion. This choice is made because the OP presents their full opinion in their original post, and a challenger typically presents their central argument in their initial challenge~\cite{tan2016winning}.
While many CMV studies predict persuasion outcomes, prediction tasks do not reveal the actual bidirectional relationship between syntactic similarity and persuasion. We use FastKASSIM to analyze this relationship.

We find that arguments which eventually lead to deltas ($\mu=0.307$) tend to be more syntactically similar to original opinions than unsuccessful ($\mu=0.263$) arguments ($t=19.016$; $p$ $<0.001$).
However, this finding does not imply on its own that \textit{syntactically similar arguments are more persuasive}.
Thus, we also examined the converse by computing the persuasion rates (proportion of threads receiving deltas) of the most syntactically similar and dissimilar arguments. We computed the syntactic similarity of each pairing of initial arguments and original opinions and grouped them into the top and bottom $33\%$ of syntactic similarity. This resulted in a minimum syntactic similarity value of $0.341$ for the top $33\%$ ($\mu=0.453$) and maximum of  $0.171$ for the bottom $33\%$ ($\mu=0.096$). We found that threads grouped in the bottom $33\%$ of syntactic similarity only had a persuasion rate of $6.377\%$, while the persuasion rate for threads grouped in the top $33\%$ was nearly twice that, $12.347\%$ ($t=19.135$; $p$ $<0.001$). These findings support the hypothesis that similar syntactic patterns play a role in persuasion --- may be an indication of stylistic familiarity for the OP.
\vspace{-2mm}
\subsection{Authorship Attribution}
\begin{table}[]
\centering
\scalebox{0.9}{
\begin{tabular}{l|l|l}
\textbf{Features} & \textbf{Acc.$_{(\sigma)}$} & \textbf{F1}$_{(\sigma)}$\\ \hline
Majority Baseline             & 0.767          & 0.868  \\
Bag of Words          & 0.892$_{(0.02)}$          & 0.867$_{(0.02)}$ \\
Bag of Words + Syntax      & 0.923$_{(0.02)}$          & 0.922$_{(0.01)}$ \\
RoBERTa & 0.939$_{(0.01)}$ & 0.935$_{(0.00)}$ \\
RoBERTa + Syntax & \textbf{0.945$_{(0.01)}$} & \textbf{0.938$_{(0.01)}$}
\end{tabular}
}
\caption{\textbf{Judgment} test set results comparing accuracy and weighted F1 score between unigram counts and unigram counts augmented with syntactic features. Standard deviation ($\sigma$) given in subscripts.}.
\label{authorship_attribution}
\vspace{-4mm}
\end{table}
Much work has examined methods for attributing authorship based upon linguistic features \cite{juola2008authorship, raghavan2010authorship, seroussi2011authorship}. The \textbf{Judgment dataset} \cite{seroussi2011ghosts} contained English judgments delivered by judges on the Australian High Court from 1913 to 1975. We classified whether 924 judgments were written by Sir Edward McTiernan or Sir George Rich during non-overlapping time periods (Rich's judgments from 1913-1928 and McTiernan's from 1965-2971). We follow the experimental design and preprocessing steps in \citet{seroussi2011authorship}\footnote{All experiments were computed on one RTX A6000 GPU.}.

To capture semantics, one setting used normalized Bag of Words with Support Vector Machines and the other used a state-of-the-art fine-tuned RoBERTa~\cite{liu2019roberta} model\footnote{Base RoBERTa (123M parameters). We set an initial learning rate of 2e-5 and a 0.01 weight decay.}. We augmented both semantic settings with a syntactic similarity feature vector --- for each classification instance, we randomly sampled 25 posts from the training set and computed the FastKASSIM syntactic similarity between judgments written by Rich and McTiernan, respectively. The syntactic similarity features consisted of the minimum, maximum, mean, and standard deviation of these comparisons. We evaluated our classifier on a $10\%$ withheld testing set\footnote{We used 4 seeds to sample our data.}.

Table~\ref{authorship_attribution} shows that adding syntactic features to both semantic models results in gains in both accuracy and weighted F1. This is even the case when using RoBERTa; we achieve the strongest performance using RoBERTa with a weighted sum between textual and syntactic features, fine-tuned using modules from the frameworks in \citet{gu2021package, wolf2020transformers}. Syntactic similarity with reference documents may provide a strong indicator of writing style.

\vspace{-2mm}
\section{Conclusion}
We have introduced FastKASSIM, which has runtime improvements that scale significantly with document sizes and achieves better agreement with human perception of syntactic differences compared to standard syntax metrics. These improvements are possible due to our Label-based Tree Kernel, which has an improved asymptotic runtime complexity and a corrected normalization. FastKASSIM also allowed us to verify hypotheses regarding the importance of syntax both in authorship attribution and social dynamics such as persuasion. These findings motivate further applications of syntax. 

\section*{Acknowledgements}
Many thanks to Lillian Lee and Chenhao Tan for their feedback and collaboration on early versions of this work.
We are very grateful to Reihane Boghrati for  providing access to their Mechanical Turk dataset from their work on CASSIM. Thanks to Jack Hessel, Kun Qian, Yu Li, Qingyang Wu, Sky Wang, Mert Inan, and Zhiyang Xu for their helpful feedback and discussions.
\section*{Limitations}
Our work relies on a couple assumptions. Our main corpus for evaluation is the crowdsourced and human-annotated dataset from \citet{boghrati2018conversation}. As a result, our claim to better represent human perception of syntax relies on the assumption that their annotators correctly filter out responses which are not actually syntactically similar to each prompt. They had an acceptable Cohen's Kappa of $0.53$. Additionally, we only use corpora that are in English. Future work should look towards applying our general approach to other languages.

In our evaluation of FastKASSIM against LSM and CASSIM, we also evaluate its ability to correctly identify statements created from the same prompt as similar and statements created from different prompts as dissimilar (Table~\ref{metric_performance}). In this evaluation, we assume that 0.50 is an acceptable threshold for syntactic similarity and dissimilarity, because without any contextual information, one would assume that there are an equal amount of similar and dissimilar documents. Despite this, we still performed quantile adjustments for each comparison metric, uniformly distributing the scores between 0.0 and 1.0. This gives is an unfair advantage for the comparison metrics (i.e., CASSIM, LSM, and BERTScore), since ``in-the-wild'' it is impossible to obtain the eventual distribution of scores. Future work may consider methods to rebalance each of these scores, including conducting human evaluation to evaluate whether 0.50 is an acceptable threshold for syntactic similarity both before and after each metric undergoes an adjustment to the uniform distribution.

FastKASSIM is a metric for syntactic similarity between a pair of utterances or documents. However, similarity is only one dimension of syntax, which removes some granularity --- for instance, syntactic similarity cannot explain which specific productions are shared. Similarity metrics like FastKASSIM instead afford opportunities in a variety of other applications, such as syntactic coherence in language generation and verifying computational social science hypotheses. 

Additionally, parsing is still a significant bottleneck in runtime. Future work may wish to consider ways to mitigate the cost of parsing. 
One may also consider using sequential modeling to generate syntactic parse trees, or to directly model the output of FastKASSIM.

\section*{Ethical Considerations}
Our study makes use of three datasets. First is the set of prompts collected in \citet{boghrati2018conversation}, which involved anonymous participants creating fictional statements, so there is no personal information involved. Second is the publicly available r/ChangeMyView dataset collected by \citet{tan2016winning}, which consists of statements made by users behind typically anonymous aliases. Lastly is the publicly available WikiQA corpus~\cite{yang-etal-2015-wikiqa}, which does not contain identifying information.

In our r/ChangeMyView application studying the relationship between syntactic similarity and persuasion, we make the assumption that r/ChangeMyView is a community representative of online arguments. However, partially due to its anonymity, it is unknown whether r/ChangeMyView is a representative sample with diversity in location, educational background, socioeconomic status, ethnicity, and many other important factors. An ideal study should be able to control for proxies for individual traits in order to isolate the impact of syntax itself.

Generally, while most algorithms are not inherently unethical, there is often potential for abuse in their applications. The individual computations in the FastKASSIM algorithm do not have any negative implications, but it is possible to use syntactic similarity for unethical downstream tasks. For instance, because syntax is an important aspect of writing style, it is possible that users may try to adversarially uncover an anonymous author's identity. We do not condone the use of FastKASSIM for any unlawful or morally unjust activities. We do not propose any new tasks that would introduce unethical activity.

\bibliography{anthology,custom}
\bibliographystyle{acl_natbib}

\clearpage
\appendix
\label{sec:appendix}
\section{Formalizing FastKASSIM}
\label{runtime_analysis}
\begin{figure}[h!]
    \centering
    \includegraphics[scale=0.7]{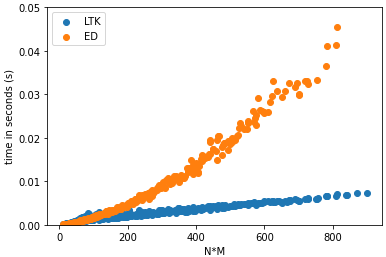}
    \caption{Runtime Comparison of Edit Distance (ED) and Label-based Tree Kernel (LTK) on WikiQA with varying $NM$ (product of parse tree sizes).}
    \label{fig:rt_nm_cmp}
\end{figure}
The Label-based Tree Kernel recursively computes $\mathrm{Delta_{lb}}$ across all $n_1, n_2$ pairs in parse trees $P_1, P_2$. 
The time complexity of $\mathrm{\Delta_{lb}}(n_1,n_2)$ has a ceiling of 
$O(L_1^{h}\times L_2^{h})$, where $L_i^{h}$ is the number of nodes at
height $h$ for a tree rooted at $n_i$, and $h$ is the minimum
height between the two trees. In the worst case scenario,
$\mathrm{\Delta_{lb}}$ of a fully uncached pair $n_i, n_j$ results in recursive calls at every depth level. $\mathrm{\Delta_{lb}}$ is computed for each node pair, so the ceiling of the tree kernel runtime is $O(NM)$, where 
$N,M$ are the total number of nodes in $P_1$ and $P_2$,
respectively.\footnote{Note that this is only possible due to $\mathrm{\Delta_{lb}}$ caching the repetitive computations when iterating over node pairs.}

$O(NM)$ is a ceiling assuming the worst-case, where the labels are the same at each comparison, requiring full recursion. Let us
consider there to be $k$ shared labels
in a pair of parse
trees. 

In each parse tree $P_1,P_2$, there will be $C_i^{(1)},C_i^{(2)}$ connected components, one for each shared label $i \in [1, k]$, where a connected component consists of connected nodes with the same label. Out of the $C_i^{(1)}$ components in parse tree $P_1$, let $N_{i,j}^{(1)}$ be the size of each individual component $j$. So, for label $i=1$, the number of comparisons follows:
\[
  O\left( \sum\limits_{l=1}^{C_1^{(1)}} \sum\limits_{m=1}^{C_1^{(2)}}
  N_{1,l}^{(1)}N_{1,m}^{(2)}  \right)
\]
which represents iterating through every pair of the $C_1^{(1)} \times C_1^{(2)}$ possible pairs of connected components and computing
LTK. Then, for $k$ shared labels, the worst-case runtime (i.e. the connected components do not form shared subtrees), we have equation~\ref{eq:connected_components}: 
\small
\begin{equation}
    O\left( \sum\limits_{i=1}^{k}  \sum\limits_{l=1}^{C_i^{(1)}}
   \sum\limits_{m=1}^{C_i^{(2)}}
 N_{i,l}^{(1)}N_{i,m}^{(2)} \right) = O\left(\sum\limits_{i=1}^{k}
 N_i^{(1)}N_i^{(2)} \right)
 \label{eq:connected_components}
\end{equation}
\normalsize
where $N_i^{(1)},N_i^{(2)}$ are the total number of nodes that have label $i$ in $P_1,P_2$ respectively. 
However, recall from Algorithm~\ref{alg:delta} that LTK only iterates through pairs that
share the same label; it does not matter if the connected components
themselves are intertwined. Then, further simplifying this term we have equation~\ref{eq:final_complexity}:
\begin{equation}
    O\left(\sum\limits_{i=1}^{k}N_i^{(1)}N_i^{(2)}\right) = O(S_{12})
    \label{eq:final_complexity}
\end{equation}
where $S_{12}$ is simply the total number of pairs in $P_1,P_2$ that have the same label.
When all nodes have the same label, $S_{12}=NM$, consistent with the observed runtime ceiling. Empirically, we see that the \textit{expectation} of $S_{12}$ is much smaller than $NM$, as seen by the sublinear time scaling in ~\autoref{fig:rt_nm_cmp}.
\subsection{Scaling with Node Pairings ($NM$)}
\begin{figure}[h!]
    \centering
    \includegraphics[scale=0.7]{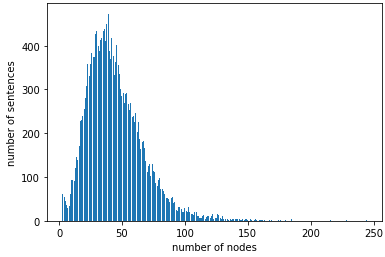}
    \caption{Statistics of Parse Trees in WikiQA}
    \label{fig:wiki_stat}
\end{figure}
We first examine the relationship between LTK and $NM$, the number of possible node pairings, using 
the WikiQA corpus~\cite{yang-etal-2015-wikiqa}, a public dataset
containing annotated question and answer pairs written in English. This dataset is chosen in order to compare FastKASSIM and CASSIM on well-structured text, and because of the ability to extract clean sentences of various length, which is crucial in determining $NM$.
The experiment utilized 20,347 answer sentences from the
WikiQA training set. We compute the parse trees prior to computing the runtimes
of Edit Distance and LTK. Statistics of parse trees from the WikiQA-train corpus are shown in ~\autoref{fig:wiki_stat}.

As the corpus is rather large, we 
consider sentences with fewer than 30 nodes, which resulted in a
total of 16,591,680 possible pairings. Then, pairings are grouped
by the product of their nodes $N \times M$. For each value of $N \times M$, we track the cost of computing both the Edit Distance and LTK for all pairings if there are less than 10 pairs, or
randomly sample 10 pairs if there are more. The average runtime for both Edit Distance and LTK for each value of $N \times M$ are shown in \autoref{fig:rt_nm_cmp}.

\section{Metrics for Evaluating FastKASSIM, CASSIM, and LSM}
\label{metric_equations}
\begin{table}[]
\centering
\begin{tabular}{l|l|l}
Metric & Sim. & Dis. \\ \hline
LSM & 70.4     & 56.0     \\
LSM$_a$ & 72.2     & 42.6     \\
CASSIM & 82.1     & 82.0     \\
CASSIM$_a$ & 48.3     & 50.6     \\
BERTScore & 89.9 & 85.1 \\
BERTScore$_a$ & 85.7 & 38.0 \\
Sentence-BERT & 60.6 & 81.4 \\ 
Sentence-BERT$_a$ & 26.7 & 57.7 \\
FastKASSIM & 73.1     & 31.7     \\ 
FastKASSIM$_a$ & 86.1     & 38.0     \\ 
\end{tabular}
\caption{Average score assigned to similarity document pairings (Sim.) and dissimilar document pairings (Dis.) by each metric. Metric$_a$ denotes an adjustment to a uniform distribution using a quantile transformation.}
\label{metric_averages}
\end{table}
In Section~\ref{discrepancies} and Table~\ref{metric_performance}, we evaluated FastKASSIM, CASSIM, and LSM in terms of Similarity Accuracy, Similar Document Recall, Similar Document Precision, Dissimilar Document Recall, and Dissimilar Document Precision. These all follow the standard formulas for accuracy, recall, and precision. Adapted to our similarity context: 

Similarity accuracy is the sum of the number of same prompt pairs receiving a score greater than 0.50 and the number of different prompt pairs receiving a score lower than 0.50 divided by the total number of pairings.

Similar document recall is the number same prompt pairs receiving a score greater than 0.50 divided by the total number of pairings originating from the same prompt.

Similar document precision is the number of same prompt pairs receiving a score greater than 0.50 divided by the total number of pairings receiving a score greater than 0.50.

Dissimilar document recall is the number different prompt pairs receiving a score less than 0.50 divided by the total number of pairings originating from different prompt.

Dissimilar document precision is the number of different prompt pairs receiving a score less than 0.50 divided by the total number of pairings receiving a score less than 0.50.

\section{Persuasiveness of Syntactic Similarity: Additional Context}
\label{sec:persuasion_appendix}
\subsection{CMV Background}
\label{sec:cmv_backgrond_appendix}
\begin{figure}[t]
    \centering
    \includegraphics[width=0.9\linewidth]{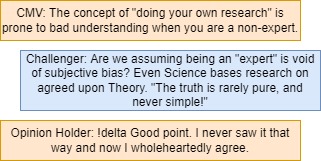}
    \caption{An example of a thread on CMV. The OP presents a set of arguments defending their opinion (orange, top), inviting challengers to contest their opinion (blue). The OP acknowledged their opinion has been changed by assigning a delta (orange, bottom).}
    \label{fig:cmv_example}
\end{figure}

We investigate the role of syntax in persuasive arguments in the r/ChangeMyView\footnote{https://www.reddit.com/r/changemyview/} community (CMV) on Reddit. CMV users come in "good faith" that they are open to changing their view on a controversial topic. They write an original post describing an opinion and allow ``challengers'' to comment on their post and attempt to change their opinion. If their opinion is changed, the original poster (OP) will indicate this by assigning a ``delta'' (by typing either ``!delta'' or $\Delta$ in response to the persuasive comment). An OP may choose to present a rebuttal to a challenger, openly disagree with a challenger, or simply ignore a challenger (e.g., Figure~\ref{fig:cmv_example}). All Reddit users use anonymous aliases, unless they explicitly disclose their identity.


Earlier work found positive relationships between \textit{behavioral mimicry} (mirroring behaviors) and in-person \textit{negotiations}~\cite{curhan2007thin,maddux2008chameleons}. Yet, \citet{healey2014divergence} found that in general spoken conversations, peoples' syntactic patterns diverged from each other. We thus investigate the hypothesis that as a challenger on CMV continues to \textit{engage} in an argument with an OP, their syntactic communication styles may begin to converge in order to ``optimize for social differences.'' Additionally, we hypothesize that challengers who utilize similar syntactic patterns, whether intentionally or not, may be more persuasive.  
\subsection{Dataset}
\label{sec:dataset}
We use the CMV dataset consisting of 18,363 posts and 1,114,533 comments written in English and collected by \citet{tan2016winning}. As in \citet{tan2016winning}, we examine discussion trees with at least 10 replies from challengers and at least one OP reply, in order to focus on discussions with ``non-trivial'' amounts of engagement. We also filter out posts which receive more than 10,000 comments in order to reduce noise from ``outsiders'' in cases where a post goes viral.
When a challenger comments on an original post, it starts a ``thread,'' with the original post (OP's opinion) taking on the root, index 0, and the challenger's comment taking on index 1. Each additional comment made in reply extends the thread. We are interested in syntactic accommodation, so we only consider threads that consist of conversations between the OP and a single challenger to eliminate confounders. These preprocessing steps results in a final dataset consisting of $15,986$ posts.
\subsection{Dicusssion \& Implications}
In Section~\ref{sec:persuasion}, we found two very statistically significant relationships between syntactic similarity and persuasive arguments. First, arguments that eventually lead to deltas tend to be more syntactically similar to original opinions compared to arguments that do not. Second, the arguments that are the most syntactically similar to original opinions actually were nearly twice as likely to receive deltas than the least syntactically similar arguments. Altogether, this may imply that syntactically similar arguments are more persuasive. This idea is supported by the rich body of work suggesting that similarity and communicative familiarity leads to improved social and conversational outcomes~\cite{curhan2007thin,giles2016communication,kaptein2014extending,maddux2008chameleons,wetzel1982similarity}.
\section{Parse Tree Examples}
\label{parse_trees}
We visualize the constituency parse trees of several sentences taken from the corpus collected in \citet{boghrati2018conversation} using the online interface of the Berkeley Neural Parser\footnote{https://parser.kitaev.io/}, which uses the parser described in \citet{kitaev2019multilingual}. 

\begin{figure*}
    \centering
    \includegraphics[width=\linewidth]{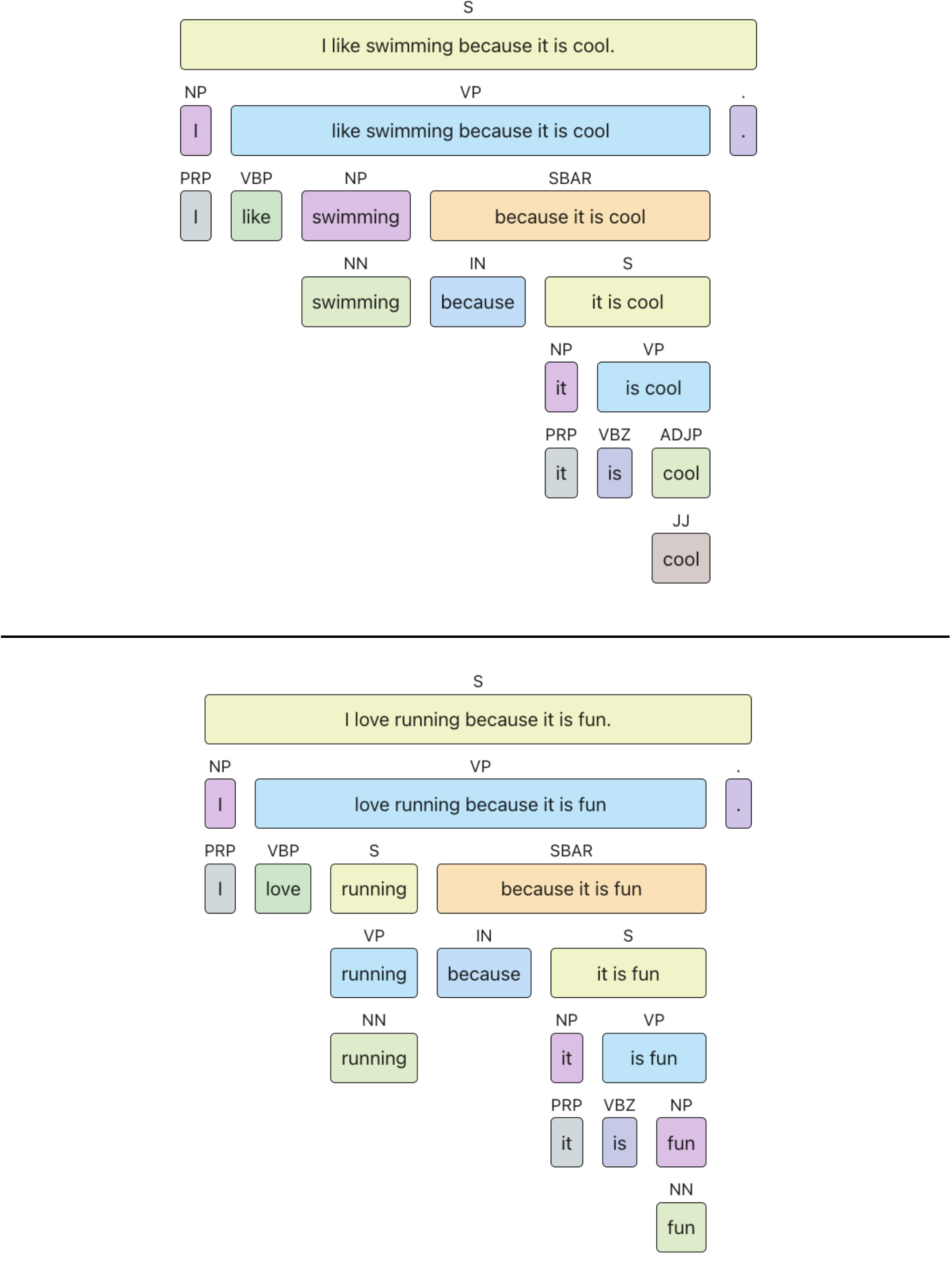}
    \caption{A comparison of the parse trees of two syntactically similar documents from \autoref{fig:metric_comparison}. Top document: ``I like swimming because it is cool.'' Bottom document: ``I love running because it is fun.'' FastKASSIM similarity score: 0.928; CASSIM similarity score: 0.962.}
    \label{fig:tree_comparison_fig1_similar}
\end{figure*}

We first compare the parse trees of the examples provided in \autoref{fig:metric_comparison}. \autoref{fig:tree_comparison_fig1_similar} compares the parse trees of the two similar documents shown in \autoref{fig:metric_comparison}. The first document is composed of one sentence --- ``I like swimming because it is cool.'' and the second document is also composed of one sentence --- ``I love running because it is fun.'' CASSIM assigned a score of 0.962, and FastKASSIM assigned a score of 0.928. The structure and composition of these two documents are nearly identical; the only difference is the production associated with the words ``running'' and ``swimming.''

\begin{landscape}
\begin{figure}[!h]
    \centering
    \scalebox{0.9}{\includegraphics[height=0.95\textheight]{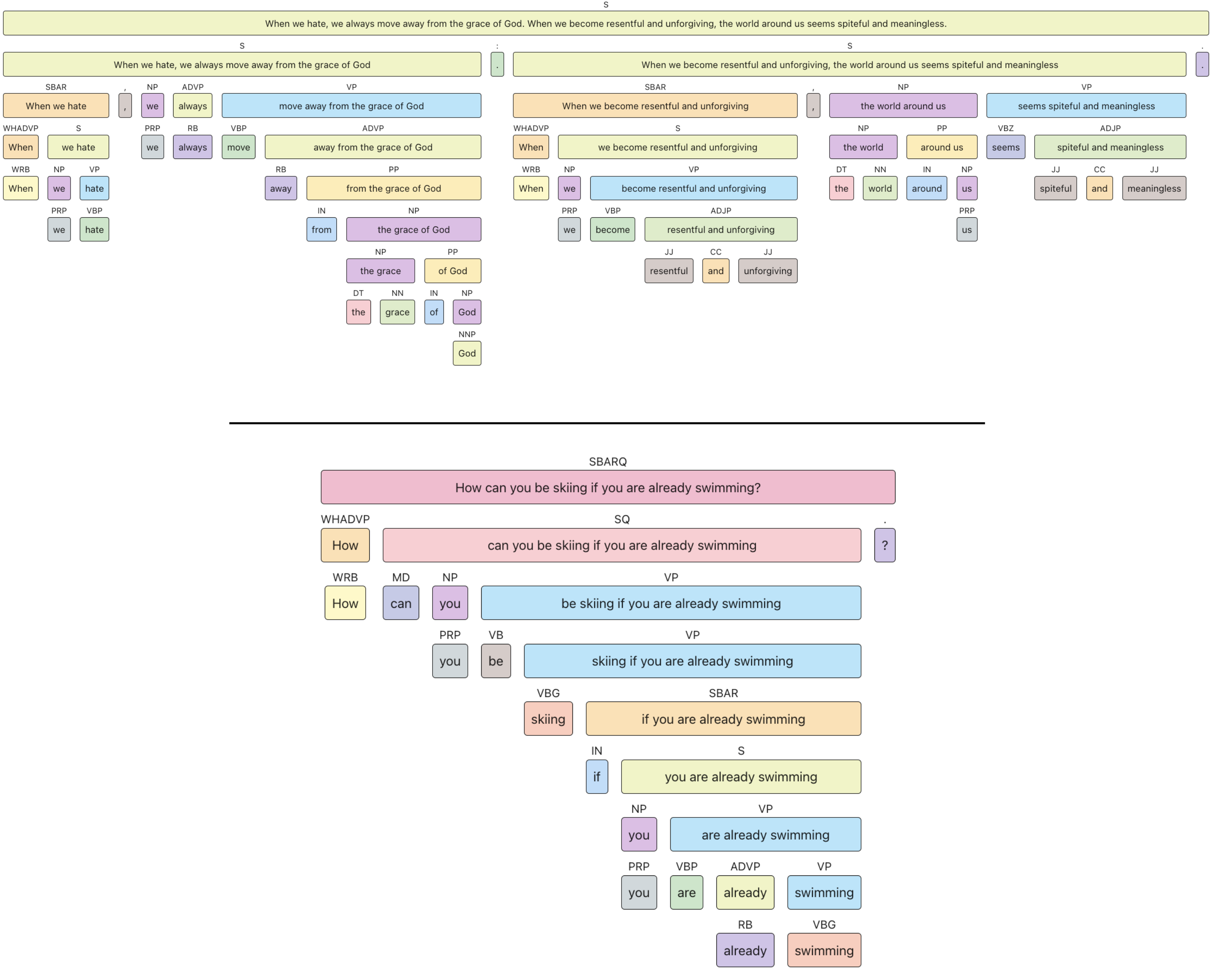}}
    \caption{A comparison of the parse trees of two syntactically dissimilar documents. Top document: ``When we hate, we always move away from the grace of God. When we become resentful and unforgiving, the world around us seems spiteful and meaningless.'' Bottom document: ``How can you be skiing if you are already swimming?'' FastKASSIM similarity score: 0.219; CASSIM similarity score: 0.838.}
    \label{fig:tree_comparison_fig1}
\end{figure}
\end{landscape}

\autoref{fig:tree_comparison_fig1} is a visualization of the parse trees of the two dissimilar documents shown in \autoref{fig:metric_comparison}. The first document is composed of two sentences: ``When we hate, we always move away from the grace of God. When we become resentful and unforgiving, the world around us seems spiteful and meaningless.'' The second document is composed of one sentence: ``How can you be skiing if you are already swimming?'' Beyond the differing number of sentences, the sentences in the first document individually appear structurally dissimilar compared to the sentence in the second document. FastKASSIM assigned a low score --- 0.219, whereas CASSIM assigned a high score --- 0.838.

\begin{figure*}
    \centering
    \includegraphics[width=\linewidth]{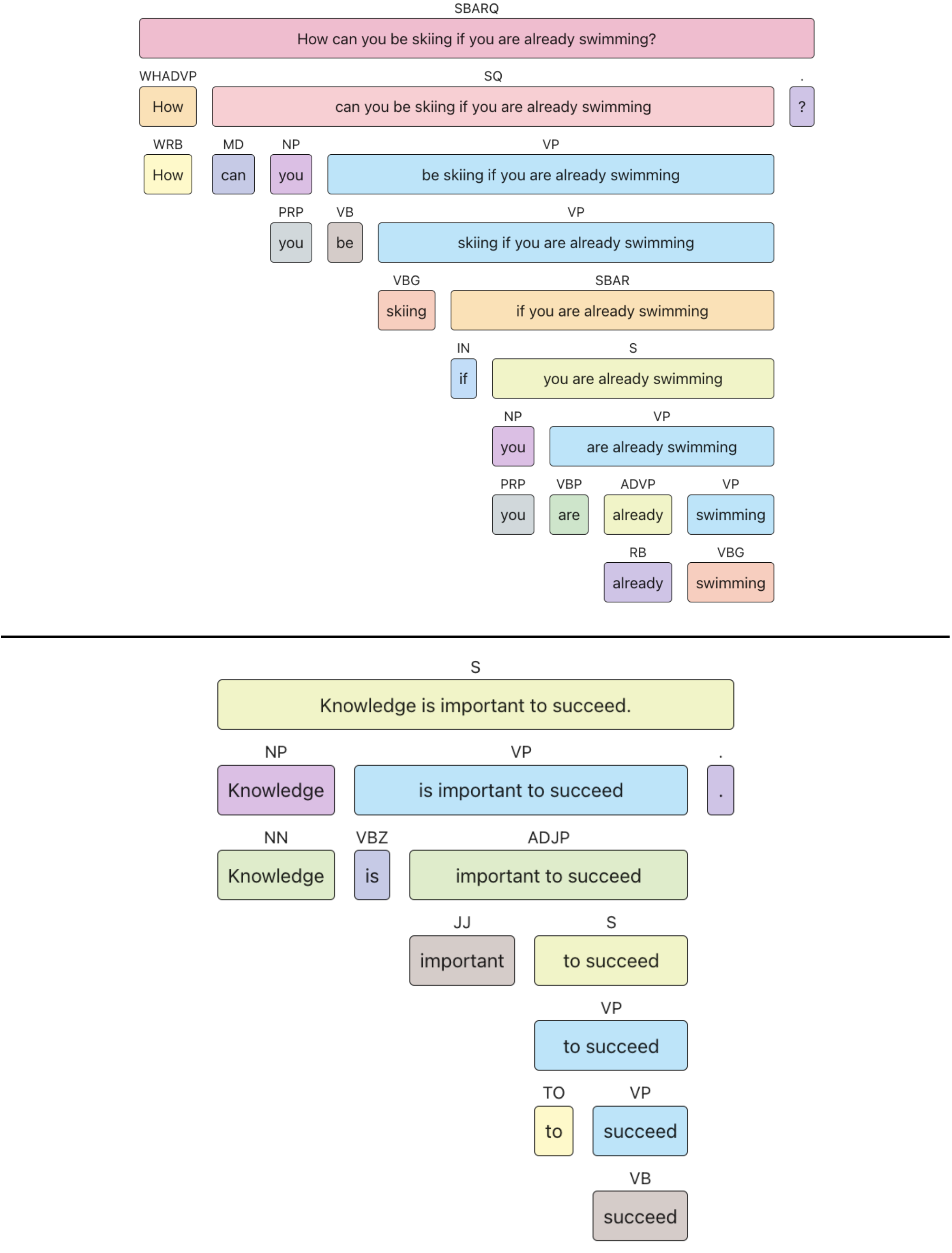}
    \caption{A comparison of the parse trees of two syntactically dissimilar documents. Top document: ``How can you be skiing if you are already swimming?'' Bottom document: ``Knowledge is important to succeed.'' FastKASSIM similarity score: 0.439; CASSIM similarity score: 0.679.}
    \label{fig:tree_comparison_1}
\end{figure*}

\autoref{fig:tree_comparison_1} compares the parse trees of the two single-sentence documents ``How can you be skiing if you already swimming?'' and ``Knowledge is important to succeed.'' As is clear from \autoref{fig:tree_comparison_1}, the two sentences are structurally and compositionally quite different. FastKASSIM assigned a score of 0.439, whereas CASSIM assigned a score of 0.679.

\begin{landscape}
\begin{figure}[!h]
   \centering
    \scalebox{0.9}{\includegraphics[height=0.95\textheight]{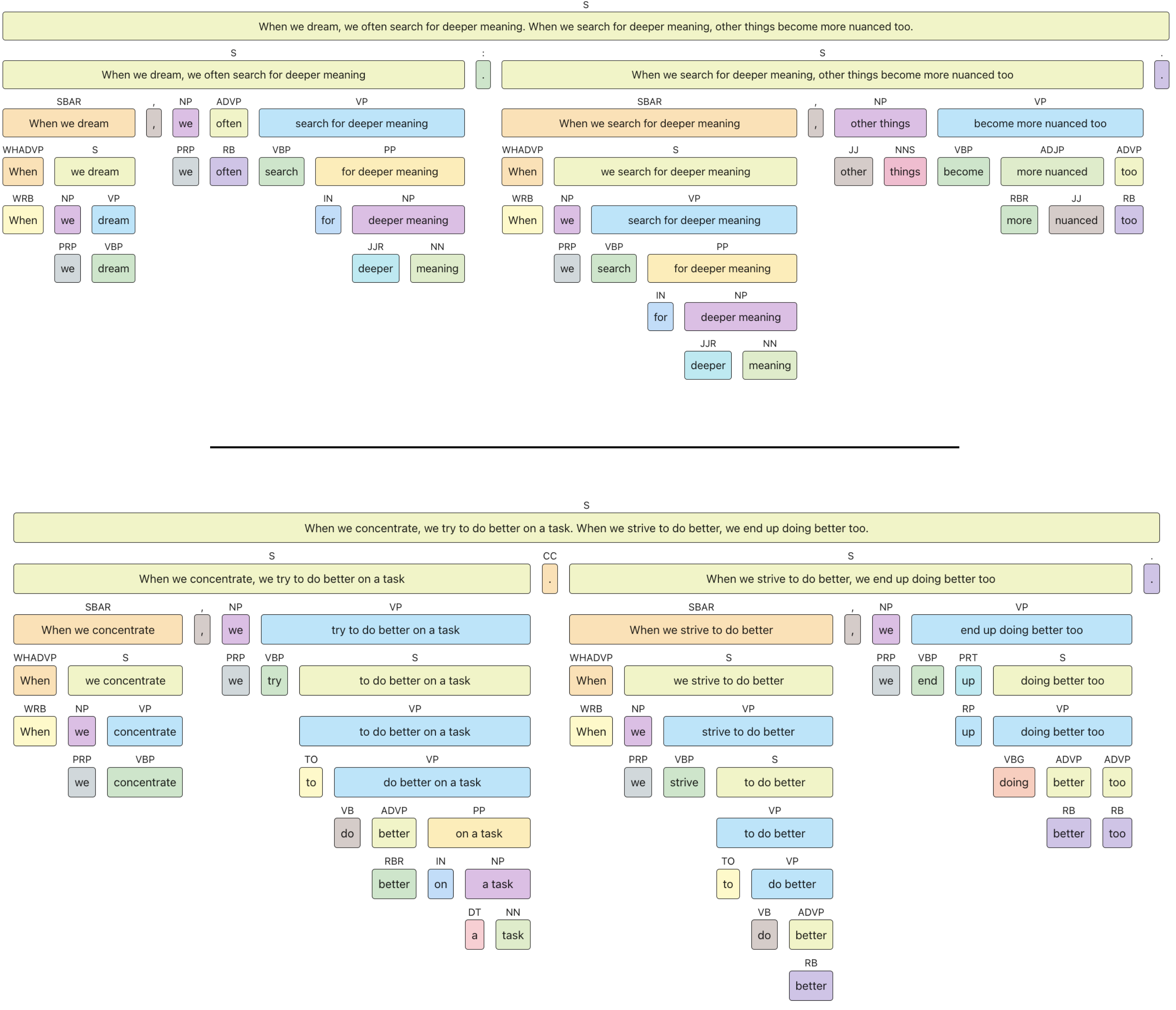}}
    \caption{A comparison of the parse trees of two syntactically similar documents. Top document: ``When we dream, we often search for deeper meaning.  When we search for deeper meaning, other things become more nuanced too.'' Bottom document: ``When we concentrate, we try to do better on a task. When we strive to do better, we end up doing better too.'' FastKASSIM similarity score: 0.656; CASSIM similarity score: 0.837.}
    \label{fig:tree_comparison_2}
\end{figure}
\end{landscape}

\autoref{fig:tree_comparison_2} compares the parse trees of two separate documents. The first document is composed of two sentences: ``When we dream, we often search for deeper meaning.  When we search for deeper meaning, other things become more nuanced too.'' The second document is composed of two sentences as well: ``When we concentrate, we try to do better on a task. When we strive to do better, we end up doing better too.'' The structures of the two documents appear rather similar, but there do appear to be some differences in composition (i.e., in terms of the constituent parts-of-speech). FastKASSIM assigned a score of 0.656, and CASSIM assigned a score of 0.837. While both scores are relatively high, FastKASSIM may be more penalizing towards these types of differences.

\begin{landscape}
\begin{figure}[!h]
   \centering
    \scalebox{0.9}{\includegraphics[height=0.95\textheight]{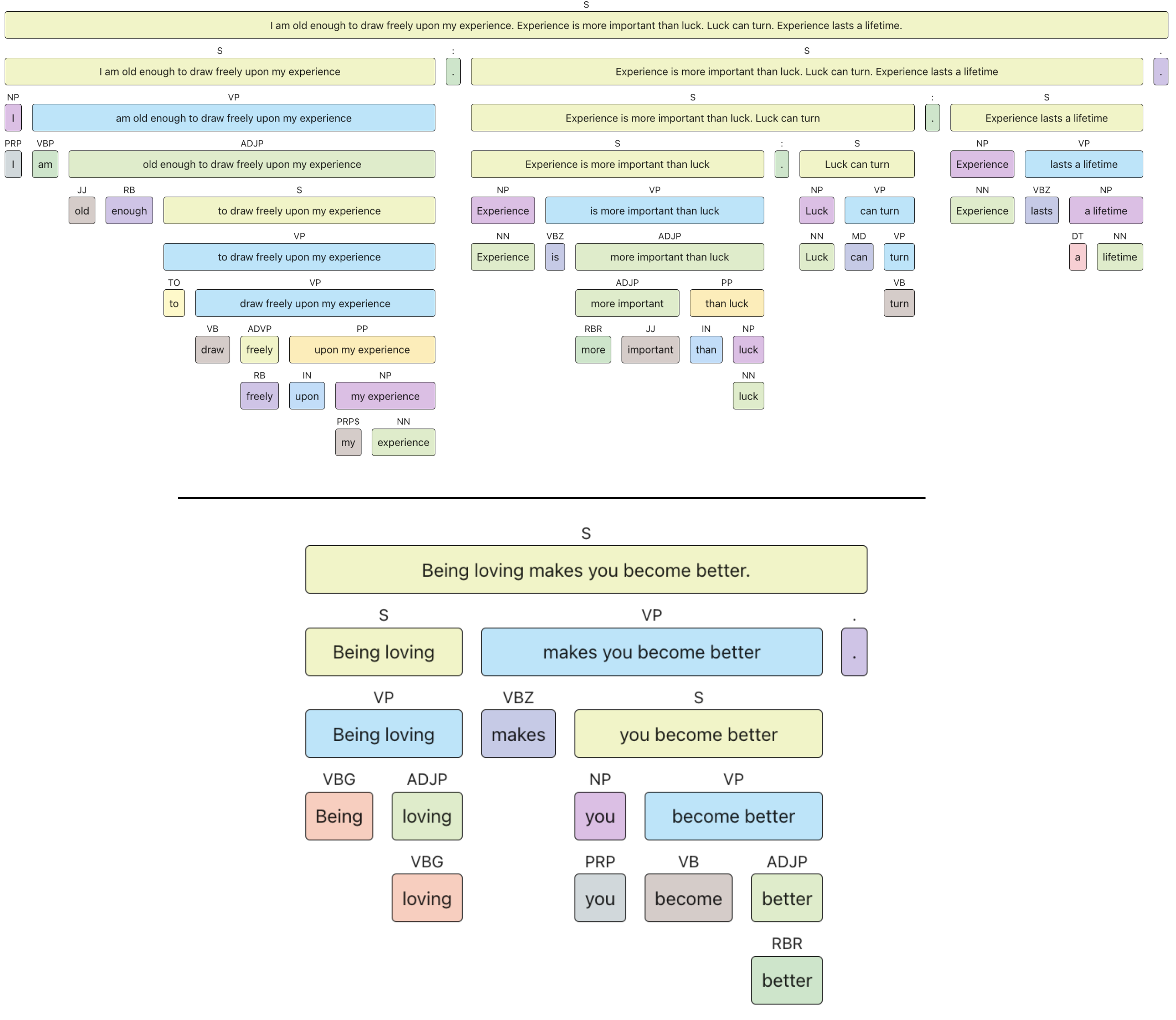}}
    \caption{A comparison of the parse trees of two syntactically dissimilar documents. Top document: ``I am old enough to draw freely upon my experience. Experience is more important than luck. Luck can turn. Experience lasts a lifetime.'' Bottom document: ``Being loving makes you become better.'' FastKASSIM similarity score: 0.15; CASSIM similarity score: 0.924.}
    \label{fig:tree_comparison_3}
\end{figure}
\end{landscape}

\autoref{fig:tree_comparison_3} compares the parse trees of two separate documents. The first document is composed of four sentences: ``I am old enough to draw freely upon my experience. Experience is more important than luck. Luck can turn. Experience lasts a lifetime.'' The second document is composed of one sentence: ``Being loving makes you become better.'' Holistically, the structures of the two documents are quite different. Beyond the differing number of sentences in each document, there are also not any individual sentences between the two documents that appear particularly syntactically similar. FastKASSIM assigned a score of 0.15, whereas CASSIM assigned a score of 0.924.

\end{document}